\pdfoutput=1
\documentclass[journal]{IEEEtran}

\usepackage[pdftex]{graphicx}

\usepackage{mdwmath}
\usepackage{mdwtab}
\usepackage{url}
\usepackage{hyperref}
\usepackage{amsmath,amsfonts,amssymb}
\usepackage{multirow} 
\usepackage{cite}
\usepackage{threeparttable}
\usepackage{adjustbox}

\newcommand{\eg}{e.\,g.,}
\newcommand{\ie}{i.\,e.,}
\newcommand{\etal}{\emph{et al.}}

\begin{document}
\title{Dense Depth Estimation in Monocular Endoscopy with Self-supervised Learning Methods}

\author{
    \IEEEauthorblockN{Xingtong~Li, Ayushi~Sinha, Masaru~Ishii, Gregory~D.~Hager,~\IEEEmembership{Fellow,~IEEE,} Austin~Reiter, Russell~H.~Taylor,~\IEEEmembership{Fellow,~IEEE,} and Mathias~Unberath}%
    \thanks{Manuscript received Feburary 20, 2019; revised September 21, 2019; accepted October 25, 2019. This work was funded in part by NIH R01-EB015530, in part by a research contract from Galen Robotics, in part by a fellowship grant from Intuitive Surgical, and in part by Johns Hopkins University internal funds. \emph{(Corresponding author: Xingtong Liu.)}}%
    \thanks{Xingtong Liu is with the Computer Science Department, Johns Hopkins University, Baltimore, MD 21287 USA (e-mail: xliu89@jh.edu).}
    \thanks{Ayushi Sinha was with the Computer Science Department, Johns Hopkins University, Baltimore, MD 21287 USA. She is now with Philips Research, Cambridge, MA 02141 USA (e-mail: asinha8@jhu.edu).}
    \thanks{Masaru Ishii is with Johns Hopkins Medical Institutions, Baltimore, MD 21224 USA (e-mail: mishii3@jhmi.edu).}
    \thanks{Gregory D. Hager is with the Computer Science Department, Johns Hopkins University, Baltimore, MD 21287 USA (e-mail: hager@cs.jhu.edu).}
    \thanks{Austin Reiter was with the Computer Science Department, Johns Hopkins University, Baltimore, MD 21287 USA. He is now with Facebook, New York, NY 10003 USA (e-mail: areiter@cs.jhu.edu).}
    \thanks{Russell H. Taylor is with the Computer Science Department, Johns Hopkins University, Baltimore, MD 21287 USA. He is a paid consultant to and owns equity in Galen Robotics, Inc. These arrangements have been reviewed and approved by JHU in accordance with its conflict of interest policy (e-mail: rht@jhu.edu).}
    \thanks{Mathias Unberath is with the Computer Science Department, Johns Hopkins University, Baltimore, MD 21287 USA (e-mail: unberath@jhu.edu).}
    }

\markboth{IEEE Transactions on Medical Imaging,~Vol.~xx, No.~xx,~xx~2019}{Liu \MakeLowercase{\textit{et al.}}: Dense Depth Estimation in Monocular Endoscopy with Self-supervised Learning Methods}

\maketitle

\begin{abstract}
	We present a self-supervised approach to training convolutional neural networks for dense depth estimation from monocular endoscopy data without \textit{a priori} modeling of anatomy or shading. Our method only requires monocular endoscopic videos and a multi-view stereo method, \eg~structure from motion, to supervise learning in a sparse manner. Consequently, our method requires neither manual labeling nor patient computed tomography (CT) scan in the training and application phases. In a cross-patient experiment using CT scans as groundtruth, the proposed method achieved submillimeter mean residual error. In a comparison study to recent self-supervised depth estimation methods designed for natural video on \emph{in vivo} sinus endoscopy data, we demonstrate that the proposed approach outperforms the previous methods by a large margin. The source code for this work is publicly available online at \url{https://github.com/lppllppl920/EndoscopyDepthEstimation-Pytorch}.
\end{abstract}


\IEEEpeerreviewmaketitle
\section{Introduction} \label{introduction}
\IEEEPARstart{M}{inimally} invasive procedures in the head and neck, \eg~functional endoscopic sinus surgery, typically employ surgical navigation systems to provide surgeons with additional anatomical and positional information. This helps them avoid critical structures, such as the brain, eyes, and major arteries, that are spatially close to the sinus cavities and must not be disturbed during surgery. Computer vision-based navigation systems that rely on the intra-operative endoscopic video stream and do not introduce additional hardware are both easy to integrate into clinical workflow and cost-effective. Such systems generally require registration of pre-operative data, such as CT scans or statistical models, to the intra-operative video data~\cite{leonard2018evaluation,sinha2018endoscopic,suenaga2015vision,yang2015vision}. This registration must be highly accurate to guarantee the reliable performance of the navigation system. To enable an accurate registration, a feature-based video-CT registration algorithm requires accurate and sufficiently dense intra-operative 3D reconstructions of the anatomy from endoscopic videos. Obtaining such reconstructions is not trivial due to problems such as specular reflectance, lack of photometric constancy across frames, tissue deformation, and so on. 

\subsection{Contributions}
In this paper, we build upon our prior work~\cite{liu2018self} and present a self-supervised learning approach for single-frame dense depth estimation in monocular endoscopy. Our contributions are as follows: (1) To the best of our knowledge, this is the \emph{first} deep learning-based dense depth estimation method that only requires monocular endoscopic images during both training and application phases. In particular, it neither needs any manual data labeling, scaling, nor any other imaging modalities such as CT. (2) We propose several novel network loss functions and layers that exploit information from traditional multi-view stereo methods and enforce geometric relationships between video frames without the requirement of photometric constancy. (3) We demonstrate that our method generalizes well across different patients and endoscope cameras.

\subsection{Related work}
Several methods have been explored for depth estimation in endoscopy. These can be grouped into traditional multi-view stereo algorithms and fully supervised learning-based methods.

Multi-view stereo methods, such as Structure from Motion (SfM)~\cite{leonard2018evaluation} and Simultaneous Localization and Mapping (SLAM)~\cite{grasa2014visual}, are able to simultaneously reconstruct 3D structure while estimating camera poses in feature-rich scenes. However, the paucity of features in endoscopic images of anatomy can cause these methods to produce sparse and unevenly distributed reconstructions. This shortcoming, in turn, can lead to inaccurate registrations. Mahmoud \etal~propose a quasi-dense SLAM-based method that explores local information around sparse reconstructions from a state-of-the-art SLAM system~\cite{mahmoud2017slam}. This method densifies the sparse reconstructions from a classic SLAM system and is also reasonably accurate. However, this approach is potentially sensitive to hyper-parameters because of the normalized cross-correlation-based matching of image patches.

Convolutional neural networks (CNN) have shown promising results in high-complexity problems including general scene depth estimation~\cite{laina2016deeper}, which benefits from local and global context information and multi-level representations. However, using CNN in a fully supervised fashion in endoscopic videos is challenging because dense ground truth depth maps that correspond directly to the real endoscopic images are hard to obtain. There are several simulation-based works that try to solve this challenge by training on synthetic dense depth maps generated from patient-specific CT data. Visentini-Scarzanella \etal~use untextured endoscopy video simulations from CT data to train a fully supervised depth estimation network and rely on another transcoder network to convert real video frames to texture independent ones required for depth prediction~\cite{visentini2017deep}. This method requires per-endoscope photometric calibration and complex registration designed for narrow tube-like structures. In addition, it remains unclear whether this method will work on in-vivo images since validation is limited to two lung nodule phantoms. Mahmood \etal~simulate pairs of color images and dense depth maps from CT data for depth estimation network training. During the application phase, they use a Generative Adversarial Network to convert real endoscopic images to simulation-like ones and then feed them to the trained depth estimation network~\cite{mahmood2018deep}. In their work, the appearance transformer network is trained separately by simply mimicking the appearance of simulated images but without knowledge of the target task, \ie~depth estimation, which can lead to decreased performance up to incorrect depth estimates. Besides simulation-based methods, hardware-based solutions exist that may be advantageous in the sense that they usually do not rely on pre-operative imaging modalities~\cite{yang2016compact,simi2013magnetically}. However, incorporating depth or stereo cameras into endoscopes is challenging and, even if possible, these cameras may still fail to acquire dense and accurate enough depth maps from endoscopic scenes for fully-supervised training because of the non-Lambertian reflectance properties of tissues and the paucity of features.

Several self-supervised approaches for single-frame depth estimation have been proposed in the generic field of computer vision~\cite{garg2016unsupervised,zhou2017unsupervised,yin2018geonet,mahjourian2018unsupervised}. However, based on our observations and experiments, these methods are not generally applicable to endoscopy because of several reasons. First, photometric constancy between frames assumed in their work is not available in endoscopy. The camera and light source move jointly, and therefore, the appearance of the same anatomy can vary substantially with different camera poses, especially for regions close to the camera. Second, appearance-based warping loss suffers from gradient locality, as observed in~\cite{yin2018geonet}. This can result in network training to get trapped in bad local minima, especially for textureless regions. Compared to natural images, the overall scarcer and more homogeneous texture of tissues observed in endoscopy, \eg~sinus endoscopy and colonoscopy, makes it even more difficult for the network to obtain reliable information from photometric appearance. Moreover, estimating a global scale from monocular images is inherently ambiguous~\cite{eigen2014depth}. In natural images, the scale can be estimated using learned prior knowledge about sizes of common objects, but there are no such visual cues in endoscopy, especially for images where instruments are not present. Therefore, approaches that try to jointly estimate depths and camera poses with correct global scales are unlikely to work in endoscopy.

The first and second points above demonstrate that the recent self-supervised approaches cannot enable the network to capture long-range correlation in either spatial or temporal dimension in imaging modalities where no lighting constancy is available, \eg~endoscopy. On the other hand, traditional multi-view stereo methods, such as SfM, are capable of explicitly capturing long-range correspondences with illumination-invariant feature descriptors, \eg~Scale-Invariant Feature Transform (SIFT), and global optimization, \eg~bundle adjustment. We argue that the estimated sparse reconstructions and camera poses from SfM are valuable and should be integrated into the network training of monocular depth estimation. We propose novel network loss functions and layers that enable the integration of information from SfM and enforce the inherent geometric constraints between depth predictions of different viewpoints. Since this approach considers relative camera and scene geometry, it does not assume lighting constancy. This makes our overall design suitable for scenarios where lighting constancy cannot be guaranteed. Because of the inherent difficulty of global scale estimation of monocular camera-based methods, we elect to only estimate depth maps up to a global scale. This not only enables self-supervised learning from results of SfM, where true global scales cannot be estimated, but also makes the trained network generalizable across different patients and scope cameras, which is confirmed by our experiments. We introduce our method in terms of data preparation, network architecture, and loss design in Section~\ref{methods}. Experimental setup and results are demonstrated in Section~\ref{experiment}, where we show that our method works on unseen patients and cameras. Further, we show that our method outperforms two recent self-supervised depth estimation methods by a large margin on \emph{in vivo} sinus endoscopy data. In Section~\ref{sec:discussion} and~\ref{sec:conclusion}, we discuss the limitations of our work and future directions to explore.

\section{Methods} \label{methods}
\begin{figure*}[t]
	\centering
	\includegraphics[width=181mm]{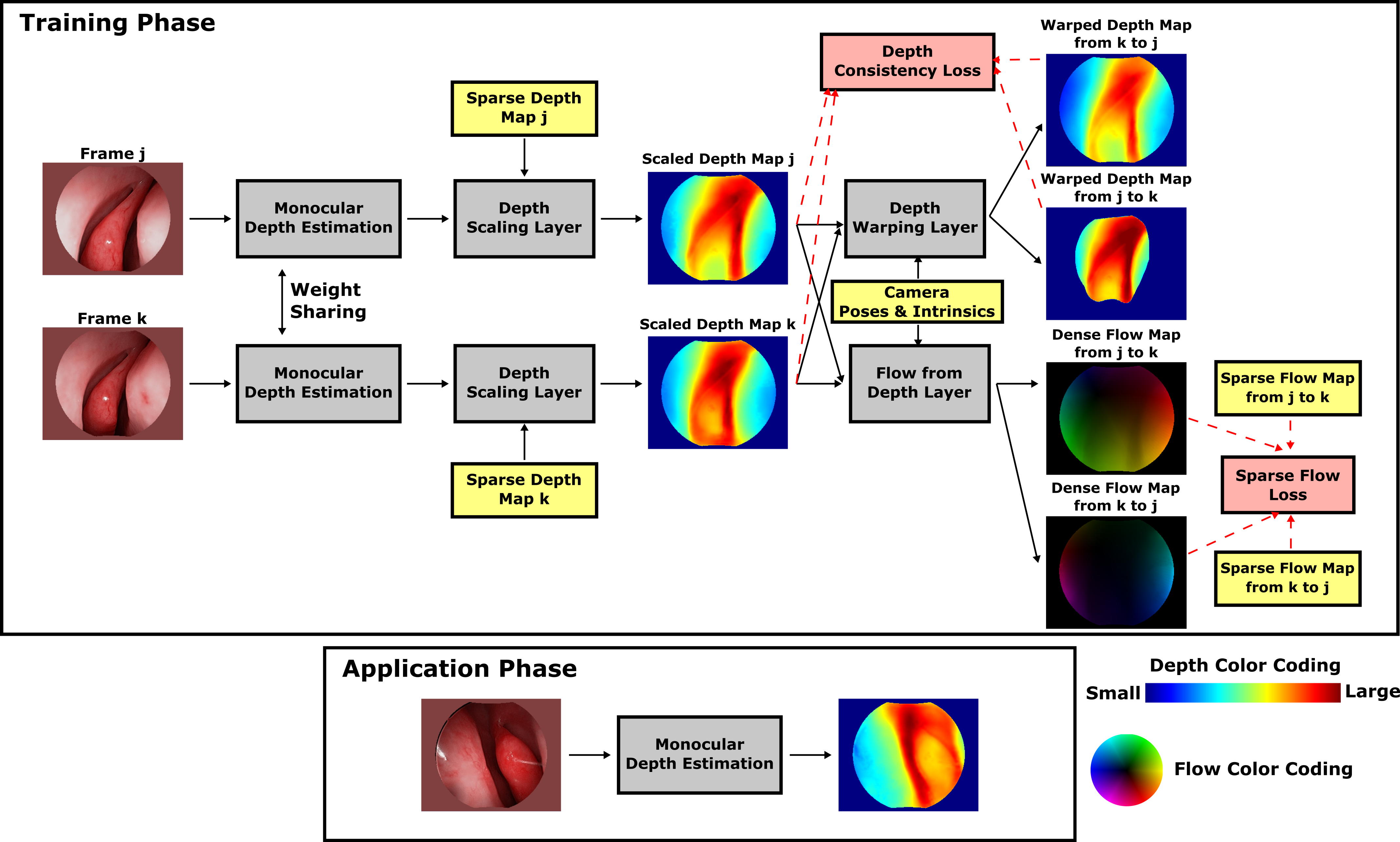} 
	\caption{\textbf{Network architecture.}\quad Our network in the training phase (top) is a self-supervised two-branch Siamese network. Two frames $j$ and $k$ are randomly selected from the same video sequence as the input to the two-branch network. To ensure enough region overlap between two frames, the frame interval is within a specified range. All concepts in the figure are introduced in Section~\ref{methods}. The red dashed arrows are used to indicate the data-loss correspondence. The warped depth map from $k$ to $j$ describes the scaled depth map $k$ viewed from the viewpoint of frame $j$. The dense flow map from $j$ to $k$ describes the 2D projection movement of the underlying 3D scene from frame $j$ to $k$. During the application phase (bottom), we use the trained weights of the single-frame depth estimation architecture, which is a modified version of the architecture in \cite{jegou2017one}, to predict a dense depth map that is accurate up to a global scale. }
	\label{fig:overall_architecture}
\end{figure*}

In this section, we describe methods to train convolutional neural networks for dense depth estimation in monocular endoscopy using sparse self-supervisory signals derived from SfM applied to video sequences. We explain how self-supervisory signals from monocular endoscopy videos are extracted, and introduce our novel network architecture and loss functions to enable network training based on these signals. The overall training architecture is shown in Fig.~\ref{fig:overall_architecture}, where all concepts are introduced in this section. Overall, the network training depends on loss functions to backpropagate useful information in the form of gradients to update network parameters. The loss functions are \emph{Sparse Flow Loss} and \emph{Depth Consistency Loss} introduced in the \hyperref[sec:lossfunctions]{Loss Functions} section. To use these two losses to guide the training of depth estimation, several types of input data are needed. The input data are endoscopic video frames, camera poses and intrinsics, sparse depth maps, sparse soft masks, and sparse flow maps, which are introduced in the \hyperref[sec:trainingdata]{Training Data} section. Finally, to convert network predictions obtained from the \emph{Monocular Depth Estimation} to proper forms for loss calculation, several custom layers are used. The custom layers are \emph{Depth Scaling Layer}, \emph{Depth Warping Layer}, and \emph{Flow from Depth Layer}, which are introduced in the \hyperref[sec:networkarchitecture]{Network Architecture} section.

\subsection {Training Data} \label{sec:trainingdata}
\begin{figure*}[t]
	\centering
	\includegraphics[width=181mm]{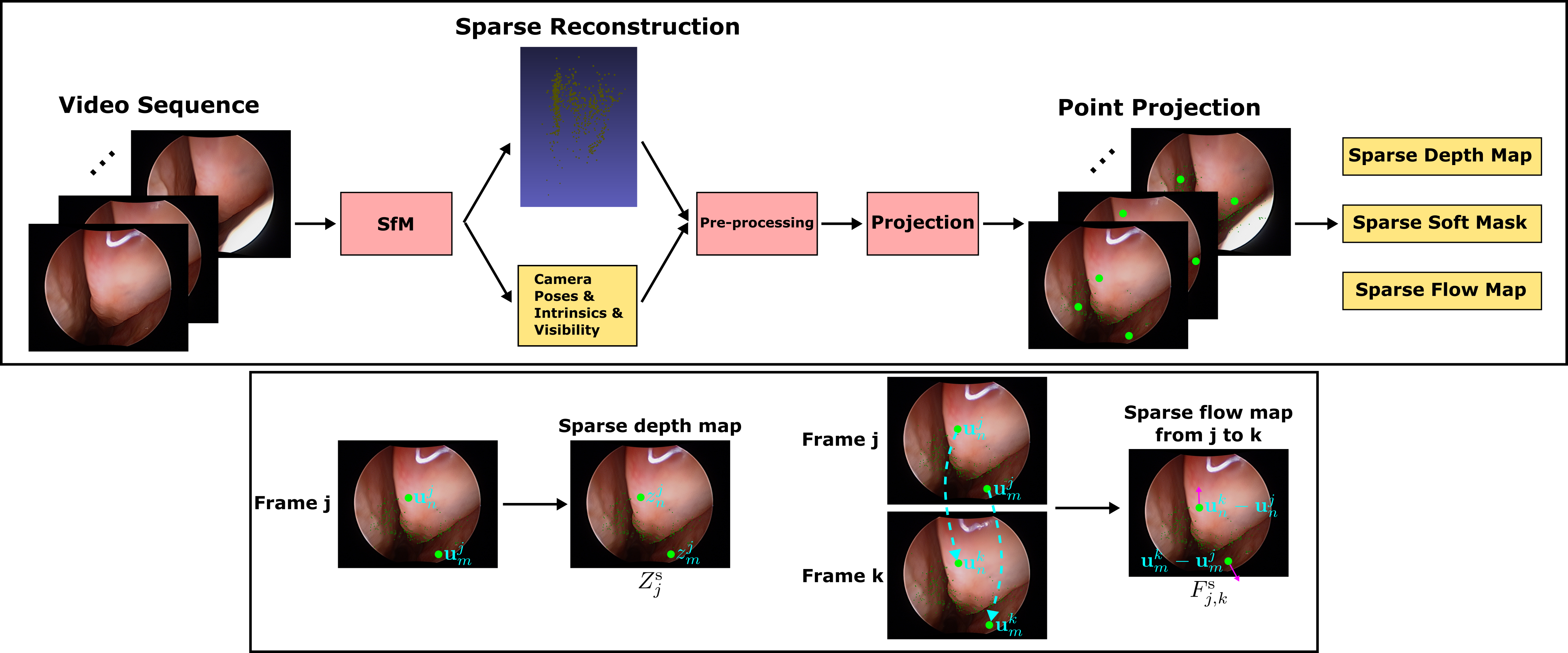}
	\caption{
	    \textbf{Training data generation pipeline.}\quad The pipeline is able to generate training data from video sequences automatically. The symbols in the figure are defined in the Training Data section. The green dots shown in the figure stand for example projected 2D locations of the sparse reconstruction. These projected 2D locations are used to store valid information for all the sparse-form data, \ie~sparse depth map, sparse soft mask, and sparse flow map. A sparse depth map stores z-axis distances of the sparse reconstruction w.r.t. the camera coordinate. A sparse soft mask stores soft weights which indicate the confidence of individual points in the sparse reconstruction. A sparse flow map stores movement of projection locations of the sparse reconstruction between two frames. The generation of a sparse depth map and sparse flow map is shown in the second row of the figure, where two example projected locations are used to demonstrate the concept. The cyan dash arrows are used to indicate point correspondences between two frames. Note that the sparse-form data do not include the color information of the videos that is used to help with the visualization of the figure.
	}
	\label{fig:training_data_generation}
\end{figure*}

 Our training data are generated from unlabeled endoscopic videos. The generation pipeline is shown in Fig.~\ref{fig:training_data_generation}. The pipeline is fully automated given endoscopic and calibration videos and could, in principle, be computed on-the-fly by replacing SfM with SLAM-based methods.

\label{sec:datapreprocessing}\textbf{Data Preprocessing.}\quad A video sequence is first undistorted using distortion coefficients estimated from the corresponding calibration video. A sparse reconstruction, camera poses, and the point visibility are estimated by SfM~\cite{leonard2018evaluation} from the undistorted video sequence, where black invalid regions in the video frames are ignored. To remove extreme outliers in the sparse reconstruction, point cloud filtering is applied. The point visibility information, appeared as $b$ below, is smoothed out by exploiting the continuous camera movement present in the video. The sparse-form data generated from SfM results are introduced below.

\textbf{Sparse Depth Map.}\quad Monocular depth estimation module, shown in Fig.\ref{fig:overall_architecture}, only predicts depths up to a global scale. However, to enable valid loss calculation, the scale of the depth prediction and the SfM results must match. Therefore, the sparse depth map introduced here is used as anchor to scale the depth prediction in the \hyperref[sec:depthscalinglayer]{\emph{Depth Scaling Layer}}. To generate sparse depth maps, 3D points from the sparse reconstruction from SfM are projected onto image planes with camera poses, intrinsics, and point visibility information. The camera intrinsic matrix is $K$. The camera pose of frame $j$ with respect to the world coordinate is $T^{j}_{\text{w}}$, where $\text{w}$ stands for world coordinate system. The homogeneous coordinate of $n$\textsuperscript{th} 3D point of the sparse reconstruction in the world coordinate is $\mathbf{p}_n^{\text{w}}$. Note that $n$ can be the index of any point in the sparse reconstruction. Frame indices used in the following equations, \eg~$j$ and $k$, can be any indices within the same video sequence. The difference of $j$ and $k$ is within a specified range to keep enough region overlap. The coordinate of $n$\textsuperscript{th} 3D point w.r.t. frame $j$, $\mathbf{p}_n^{j}$, is
\begin{equation}
\mathbf{p}_n^{j} = T^{j}_{\text{w}} \mathbf{p}_n^{\text{w}} \quad \text{.}
\end{equation}
The depth of $n$\textsuperscript{th} 3D point w.r.t. frame $j$, $z_n^j$, is the $\text{z}$-axis component of $\mathbf{p}_n^{j}$. The 2D projection location of $n$\textsuperscript{th} 3D point w.r.t. frame $j$, $\mathbf{u}_n^j$, is
\begin{equation}
\mathbf{u}_n^j = K \dfrac{\mathbf{p}_n^{j}}{z_n^j} \quad \text{.}
\end{equation}
We use $b_n^{j}=1$ to indicate that $n$\textsuperscript{th} 3D point is visible to frame $j$ and $b_n^{j}=0$ to indicate otherwise. Note that the point visibility information from SfM is used to assign the value to $b_n^j$. The sparse depth map of frame $j$, $Z_j^{\text{s}}$, is
\begin{equation}
    Z_j^{\text{s}}\left(\mathbf{u}_n^j\right) = 
    \begin{cases} 
    z_n^{j} \quad \text{if} \, b_n^{j}=1\\
    0 \quad \text{if} \, b_n^{j}=0
    \end{cases} \quad \text{, where}
\end{equation}
$\text{s}$ stands for word "sparse". Note that for equations in the Training Data section, they describe the value assignments for regions where points of the sparse reconstruction project onto. For regions where no points project onto, the values are set to zero.

\textbf{Sparse Flow Map.}\quad The sparse flow map is used in the \hyperref[sec:sparseflowloss]{\emph{Sparse Flow Loss}} introduced below. Previously, we directly used the sparse depth map for loss calculation~\cite{liu2018self} to exploit self-supervisory signals of sparse reconstructions. This makes the training objective, \ie~sparse depth map, for one frame fixed and potentially biased. Unlike the sparse depth map, sparse flow map describes the 2D projected movement of the sparse reconstruction, which involves camera poses of two input frames with random frame interval. By combining the camera trajectory and sparse reconstruction, and considering all pair-wise frame combinations, the error distribution of the new objective, \ie~sparse flow map, for one frame is more likely to be unbiased. This makes the network less affected by the random noise in the training data. We observe that the depth predictions are naturally smooth with edge-preserving for the model trained with SFL, which removes the need of explicit regularization during training, \eg~smoothness loss proposed in Zhou~\etal~\cite{zhou2017unsupervised} and Yin~\etal~\cite{yin2018geonet}.

The sparse flow map, $F_{j,k}^{\text{s}}$, represents the 2D projected movement of the sparse reconstruction from frame $j$ to frame $k$.
\begin{equation}
    F_{j,k}^{\text{s}}\left(\mathbf{u}_n^j\right) = 
    \begin{cases} 
    \dfrac{\mathbf{u}_n^k - \mathbf{u}_n^j}{\left(W, H\right)^\intercal} \quad \text{if} \, b_n^{j}=1\\
    \boldsymbol{0} \quad \text{if} \, b_n^{j}=0
    \end{cases}\quad  \text{, where}
\end{equation}
$H$ and $W$ are the height and width of the frame, respectively.

\textbf{Sparse Soft Mask.}\quad A sparse mask enables the network to exploit the valid sparse signals in the sparse-form data and ignore the rest of the invalid regions. The soft weighting is defined before training and accounts for the fact that the error distribution of individual points in the results of SfM is different and mitigates the effect of reconstruction errors from SfM. It is designed with the intuition that a larger number of frames used in triangulating one 3D point in the bundle adjustment of SfM usually means higher accuracy. The sparse soft mask is used in the SFL introduced below. The sparse soft mask of frame $j$, $M_j$, is defined as
\begin{equation}
    M_j\left(\mathbf{u}_n^j\right) = \\
    \begin{cases}
            1 - \mathrm{e}^{{-\sum_i b_n^{i}} / \sigma} \quad \text{if} \, b_n^{j}=1\\
            0 \quad \text{if} \, b_n^{j}=0
    \end{cases} \quad \text{, where}
\end{equation}
$i$ iterates over all frames in the video sequence where the SfM is applied. $\sigma$ is a hyper-parameter based on the average number of frames used to reconstruct each sparse point in SfM.

\subsection {Network Architecture} \label{sec:networkarchitecture}
Our overall network architecture shown in Fig.~\ref{fig:overall_architecture} consists of a two-branch Siamese network~\cite{chopra2005learning} in the training phase. It relies on sparse signals from SfM and geometric constraints between two frames to learn to predict dense depth maps from single endoscopic video frames. In the application phase, the network has a simple single-branch architecture for depth estimation from a single frame. All the custom layers below are \emph{differentiable} so that the network can be trained in an end-to-end manner.

\textbf{Monocular Depth Estimation.}\quad This module uses a modified version of the $57$-layer architecture in~\cite{jegou2017one}, known as DenseNet, which achieves comparable performance with other popular architectures with a large reduction of network parameters by extensively reusing preceding feature maps. We change the number of channels in the last convolutional layer to $1$ and replace the final activation, which is log-softmax, with linear activation to make the architecture suitable for the task of depth prediction. We also replace the transposed convolutional layers in the up transition part of the network with nearest neighbor upsampling and convolutional layers to reduce the checkerboard artifact of the final output~\cite{odena2016deconvolution}.

\label{sec:depthscalinglayer}\textbf{Depth Scaling Layer.}\quad This layer matches the scale of the depth prediction from \emph{Monocular Depth Estimation} and the corresponding SfM results for correct loss calculation. Note that all operations of the following equations are element-wise except that $\sum$ here is summation over all elements of a map. $Z^{\prime}_j$ is the depth prediction of frame $j$ that is correct up to a scale. The scaled depth prediction of frame $j$, $Z_j$, is
\begin{equation}
  Z_j = \left(\dfrac{1}{\sum M_j} \sum { \left ( M_j \dfrac{Z_j^{\text{s}}}{Z_j^{\prime} + \epsilon} \right )}\right) Z_j^{\prime} \quad \text{, where}
\end{equation}
$\epsilon$ is a hyper-parameter to avoid zero division.

\textbf{Flow from Depth Layer.}\quad To use the sparse flow map generated from SfM results to guide network training with the SFL described later, the scaled depth map first needs to be converted to a dense flow map with the relative camera poses and the intrinsic matrix. This layer is similar to the one proposed in~\cite{yin2018geonet}, where they use the produced dense flow map as the input to an optical flow estimation network. Here instead, we use it for the depth estimation training. The dense flow map is essentially a 2D displacement field describing a 3D viewpoint change. Given the scaled depth map of frame $j$, and the relative camera pose of frame $k$ w.r.t. frame $j$, $T_j^{k}=\left(R_j^{k}, \boldsymbol{t}_j^{k}\right)$, a dense flow map between frame $j$ and $k$, $F_{j,k}$, can be derived. To demonstrate the operations in a parallelizable and differentiable way, the equations below are described in a matrix form. The 2D locations in frame $j$, $\left(U, V\right)$, are organized as a regular 2D meshgrid. The corresponding 2D locations of frame $k$ are $\left(U_k, V_k\right)$, which are organized in the same spatial arrangement as frame $j$. $\left(U_k, V_k\right)$ is given by
\begin{equation}
    \begin{aligned}
       &U_k = \dfrac{Z_j \left(A_{0, 0} U + A_{0, 1} V + A_{0, 2}\right) + B_{0, 0}}{Z_j \left(A_{2, 0} U + A_{2, 1} V + A_{2, 2}\right) + B_{2, 0}} \\
       &V_k = \dfrac{Z_j \left(A_{1, 0} U + A_{1, 1} V + A_{1, 2}\right) + B_{1, 0}}{Z_j \left(A_{2, 0} U + A_{2, 1} V + A_{2, 2}\right) + B_{2, 0}}
   \end{aligned} \quad \text{.}
\end{equation}
As a regular meshgrid, $U$ consists of $H$ rows of $\left[0, 1, \dots, W-1\right]$, and $V$ consists of $W$ columns of $\left[0, 1, \dots, H-1\right]^T$. $A = K R_{j}^{k} K^{-1}$ and $B = -K \boldsymbol{t}_{j}^{k}$. $A_{m,n}$ and $B_{m,n}$ are elements of $A$ and $B$ at position $\left(m,n\right)$, respectively. The dense flow map, $F_{j,k}$, for describing the 2D displacement field from frame $j$ to frame $k$ is
\begin{equation}
    F_{j,k}=\left(\dfrac{U_k - U}{W}, \dfrac{V_k - V}{H} \right) \quad \text{.}
\end{equation}

\textbf{Depth Warping Layer.}\quad The sparse flow map mainly provides guidance to regions of a frame where sparse information from SfM gets projected onto. Given that most frames only have a small percentage of pixels whose values are valid in a sparse flow map, most regions are still not properly guided. With the camera motion and camera intrinsics, geometric constraints between two frames can be exploited by enforcing consistency between the two corresponding depth predictions. The intuition is that the dense depth maps predicted separately from two neighbor frames are correlated because there is overlap between the observed regions. To make the geometric constraints enforced in the \hyperref[sec:depthconsistencyloss]{\emph{Depth Consistency Loss}} described later differentiable, the viewpoints of the depth predictions must be aligned first. Because a dense flow map describes a 2D projected movement of the observed 3D scene, $U_k$ and $V_k$ described above can be used to change the viewpoint of the depth $Z_k$ from frame $k$ to frame $j$ with an additional step, which is modifying $Z_k$ to describe the depth value changes due to the viewpoint changing. The modified depth map of frame $k$, $\tilde{Z}_k$, is
\begin{equation}
  \tilde{Z}_k = Z_k \left(C_{2, 0} U + C_{2, 1} V + C_{2, 2}\right) + D_{2, 0} \quad \text{, where}
\end{equation} 
$C = K R_{k}^{j} K^{-1}$, $D = K \boldsymbol{t}_{k}^{j}$. With $U_k$, $V_k$ and $\tilde{Z}_k$, the bilinear sampler in~\cite{jaderberg2015spatial} is able to generate the dense depth map $\check{Z}_{k,j}$ that is warped from the viewpoint of frame $k$ to that of frame $j$

\subsection {Loss Functions} \label{sec:lossfunctions}
We propose novel losses that can exploit self-supervisory signals from SfM and enforce geometric consistency between depth predictions of two frames. 

\label{sec:sparseflowloss}\textbf{Sparse Flow Loss (SFL).}\quad To produce correct dense depth maps that agree with sparse reconstructions from SfM, the network is trained to minimize the differences between the dense flow maps and the corresponding sparse flow maps. This loss is scale-invariant because it considers the difference of the 2D projected movement in the unit of pixel, which solves the data imbalance problem caused by the arbitrary scales of SfM results. The SFL associated with frame $j$ and $k$ is calculated as

\begin{equation}
    \begin{aligned}
        \mathcal{L}_{\text{flow}}\left(j, k\right) = &\dfrac{1}{\sum M_j} \sum { \left ( M_j \lvert F_{j,k}^{\text{s}} - F_{j,k} \rvert \right )} +\\
        &\dfrac{1}{\sum M_k} \sum { \left ( M_k \lvert F_{k,j}^{\text{s}} - F_{k,j} \rvert \right )} \quad \text{.}
    \end{aligned}
\end{equation}

\label{sec:depthconsistencyloss}\textbf{Depth Consistency Loss (DCL).}\quad Sparse signals from the SFL alone could not provide enough information to enable the network to reason about regions where no sparse annotations are available. Therefore, we enforce geometric constraints between two independently predicted depth maps. The DCL associated with frame $j$ and $k$ is calculated as

\begin{equation}
\begin{aligned}
    \mathcal{L}_{\text{consist}}\left(j, k\right) = &\dfrac{\sum\left(W_{j,k} \left( Z_j - \check{Z}_{k,j} \right)^2\right)}{\sum \left(W_{j,k} \left(Z_j^2 + \check{Z}_{k,j}^2\right)\right)} +\\
    &\dfrac{\sum\left(W_{k,j} \left( Z_k - \check{Z}_{j,k} \right)^2\right)}{\sum \left(W_{k,j} \left(Z_k^2 + \check{Z}_{j,k}^2\right)\right)} \quad{,}
\end{aligned}
\end{equation}
where $W_{j,k}$ is the intersection of valid regions of $Z_j$ and the dense depth map $\check{Z}_{j,k}$ that is predicted from frame $k$ but warped to the viewpoint of frame $j$. Because SfM results contain arbitrary global scales, this loss only penalizes the relative difference between two dense depth maps to avoid data imbalance.

\textbf{Overall Loss.}\quad The overall loss function for network training with a single pair of training data from frames $j$ and $k$ is
\begin{equation}
    \mathcal{L}\left(j, k\right) = \lambda_1 \mathcal{L}_{\text{flow}}\left(j, k\right) + \lambda_2 \mathcal{L}_{\text{consist}}\left(j, k\right) \quad {.}
\end{equation}

\section{Experiment and Results} \label{experiment}

\begin{figure*}[t]
	\centering
	\includegraphics[width=181mm]{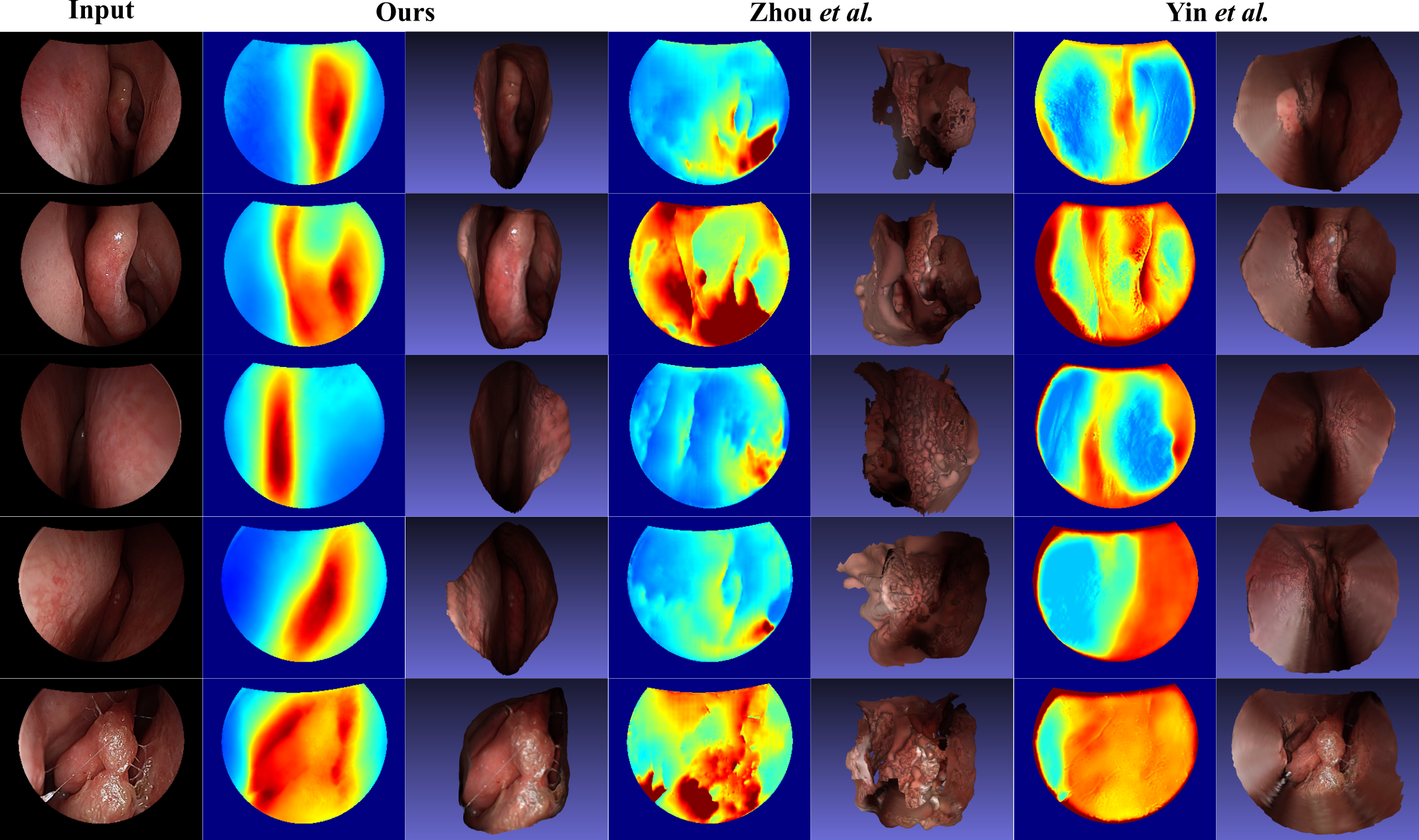} 
	\caption{
	    \textbf{Qualitative result comparison between our method, Zhou~\etal~\cite{zhou2017unsupervised}, and Yin~\etal~\cite{yin2018geonet}.}\quad The first column consists of testing and training images, where the first $3$ are testing ones. The second and third columns consist of corresponding depth maps and reconstructions from our method. The fourth and fifth columns are from Zhou~\etal. The last two columns are from Yin~\etal. For each displayed video frame, a sparse depth map is used to re-scale depth predictions from three methods. The scaled depth predictions are then normalized with the same max depth values for 2D visualization, where the same depth color coding as Fig.~\ref{fig:overall_architecture} is used. The point clouds converted from the depth predictions are post-processed by a standard Poisson surface reconstruction method~\cite{kazhdan2013screened} for 3D visualization. It shows that our method performs consistently better than Zhou~\etal and Yin~\etal in both testing and training cases.
	}
	\label{fig:result_comparison}
\end{figure*}

\subsection{Experiment Setup} 
All experiments are conducted on a workstation with $4$ NVIDIA Tesla M60 GPU, each with $8$ GB memory. The method is implemented using PyTorch~\cite{paszke2017automatic}. The dataset contains $10$ rectified sinus endoscopy videos acquired with different endoscopes. The videos were collected from $8$ anonymized and consenting patients and from $2$ cadavers under an IRB approved protocol. The overall duration of videos is approximately $30$ minutes. In all leave-one-out experiments below, the data from $7$ out of $8$ patients are used for training. The data from the $2$ cadavers are used for validation and the left-out patient is used for testing.

We select trained models for evaluation based on the network loss on the validation dataset. Overall, two types of evaluation are conducted. One is comparing point clouds converted from depth predictions with the corresponding surface models from CT data. The other is directly comparing depth predictions with the corresponding sparse depth maps generated from SfM results.

For the evaluation related to CT data, we pick $20$ frames with sufficient anatomical variation per testing patient. The depth predictions are converted to point clouds. The initial global scales and poses of point clouds before registration are manually estimated. To this end, we pick the same set of anatomical landmarks in both the point cloud and the corresponding CT surface model. $3000$ uniformly sampled points from each point cloud are registered to the corresponding surface models generated from the patient CT scans~\cite{sinha2017simultaneous} using Iterative Most Likely Oriented Point (IMLOP) algorithm~\cite{billings2014iterative}. We modify the registration algorithm to estimate a similarity transform with hard constraint during optimization. The constraint is to prevent the point cloud from deviating from the initial alignment too much, given that the initial alignments are approximately correct. The residual error is defined as the average Euclidean distance over all closest point pairs of the registered point cloud to the surface model. The average residual errors over all point clouds are used as the accuracy estimate of the depth predictions.

For the evaluation related to SfM, all video frames of the testing patient where a valid camera pose is estimated by SfM are used. Sparse depth maps are first generated from the SfM results. For a fair comparison, all depth predictions are first re-scaled with the corresponding sparse depth maps using the \emph{Depth Scaling Layer} to match the scale of the depth predictions and SfM results. Because of the scale ambiguity of the SfM results, we only use common scale-invariant metrics for evaluation. The metrics are Absolute Relative Difference, which is defined as: $\frac{1}{\lvert T \rvert}\Sigma_{y\in T} \lvert y - y^*\rvert / y^*$, and Threshold, which is defined as: \% of $y$ $\text{s.t.}$ $\text{max}\left( \dfrac{y_i}{y_i^*}, \dfrac{y_i^*}{y_i} \right) < \sigma$, with three different $\sigma$, which are $1.25$, $1.25^2$, and $1.25^3$~\cite{yin2018geonet}. The metrics are only evaluated on the valid positions in the sparse depth maps and the corresponding locations in the depth predictions.

In terms of the sparsity of the reconstructions from SfM. The number of points per sparse reconstruction is $4687 \left(\pm 6276\right)$. After smoothing out the point visibility information from SfM, the number of projected points per image from the sparse reconstruction is $1518 \left(\pm 1280\right)$. Given the downsampled image resolution, this means that $1.85 \left(\pm 1.56\right)\%$ of pixels in the sparse-form data have valid information. In the training and application phase, all images extracted from the videos are cropped to remove the invalid blank regions and downsampled to the resolution of $256\times320$. The range for smoothing the point visibility information in the Data Preprocessing section is set to $30$. The frame interval of two frames that are randomly selected from the same sequence and fed to the two-branch training network is set to $\left[5, 30\right]$. We use extensive data augmentation during experiments to make the training data distribution unbiased to specific patients or cameras as much as possible, \eg~random brightness, random contrast, random gamma, random HSV shift, Gaussian blur, motion blur, jpeg compression, and Gaussian noise. During network training, we use Stochastic Gradient Descent (SGD) optimization with momentum set to $0.9$ and cyclical learning rate scheduler~\cite{smith2017cyclical} with learning rate from $1.0\,\text{e}^{-4}$ to $1.0\,\text{e}^{-3}$. The batch size is set to $8$. The $\sigma$ for generating the soft sparse masks is set to the average track length of points in the sparse reconstructions from SfM. The $\epsilon$ in the depth scaling layer is set to $1.0\text{e}^{-8}$. We train the network with $80$ epochs in total. $\lambda_1$ is always $20.0$. For the first $20$ epochs, $\lambda_2$ is set to $0.1$ to mainly use SFL for initial convergence. For the remaining $60$ epochs, $\lambda_2$ is set to $5.0$ to add more geometric constraints to fine-tune the network.

\subsection{Cross-patient Study}\label{sec:crosspatientstudy}
\begin{figure}[t]
	\centering
	\includegraphics[width=88mm]{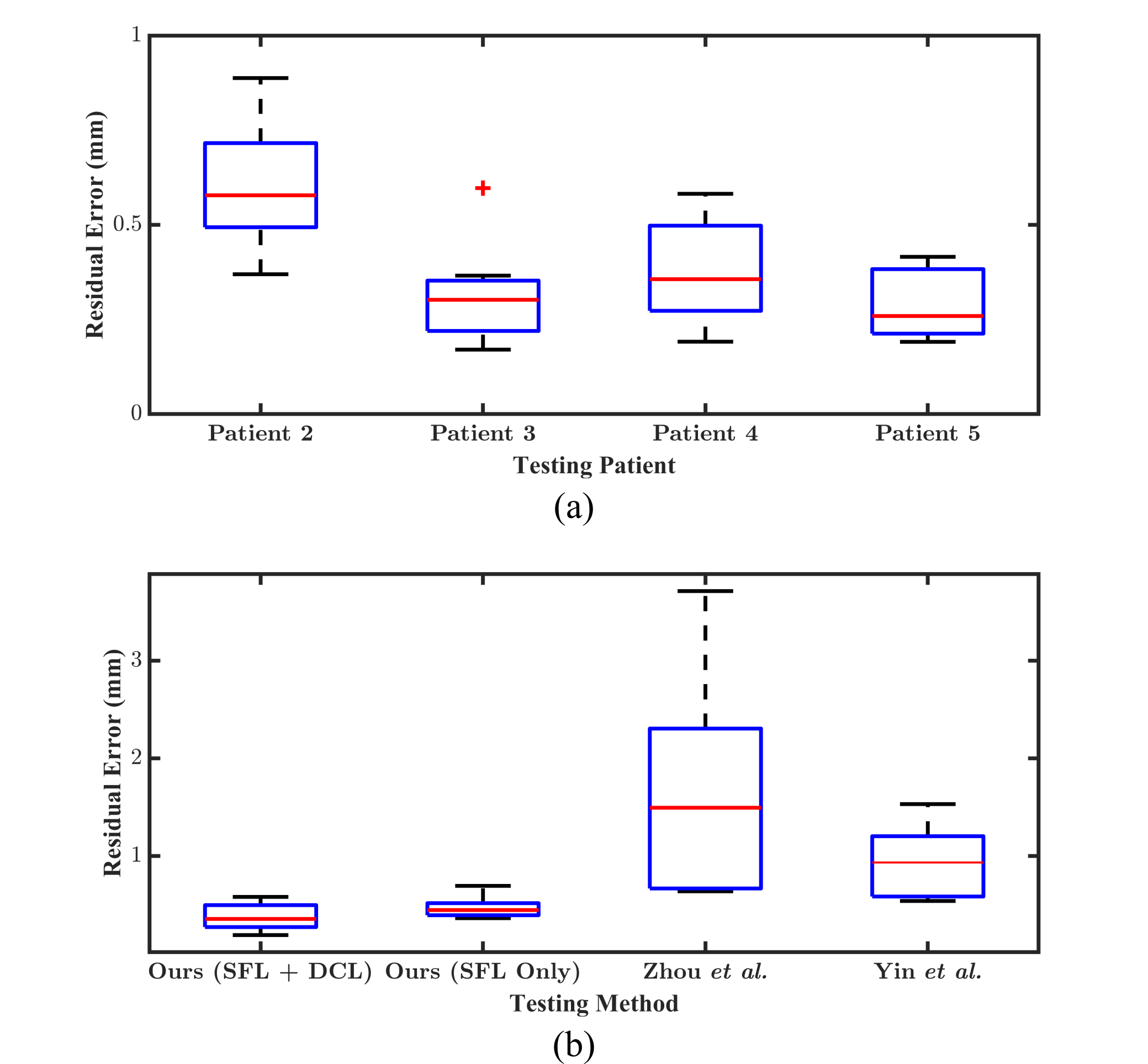}
	\caption{
	    \textbf{(a) Boxplot of residual errors for cross-patient study.}\quad The id's of the testing patients are used as labels on the horizontal axis. All testing reconstructions have submillimeter residual errors. \textbf{(b) Boxplot of residual errors for comparison study and ablation study.}\quad We compare our method with Zhou~\etal~\cite{zhou2017unsupervised} and Yin~\etal~\cite{yin2018geonet} quantitatively using data from Patient $4$ for testing. The difference between the residual errors from ours and the other two methods are statistically significant ($p<.001$). For ablation study, a model is trained with SFL only to compare with the model trained with both SFL and DCL.
	}
	\label{fig:boxplots}
\end{figure}

To show the generalizability of our method, we conduct $4$ leave-one-out experiments where we leave out Patient $2$, $3$, $4$, and $5$, respectively, during training for evaluation. Data from other patients are not used for evaluation for the lack of corresponding CT scans. The quantitative evaluation results in Fig.~\ref{fig:boxplots} (a) show that our method achieves submillimeter residual errors for all testing reconstructions. The average residual error over testing frames from all $4$ testing patients is $0.40 \left(\pm 0.18\right) \text{mm}$. For a better understanding of the accuracy of the reconstructions, the average residual error reported by Leonard~\etal~\cite{leonard2018evaluation}, where the same SfM algorithm that we use to generate training data is evaluated, is $0.32 \left(\pm 0.28\right) \text{mm}$. We use the same clinical data for evaluation as theirs in this work. Therefore, it shows our method achieves comparable performance with the SfM algorithm~\cite{leonard2018evaluation}, though our reconstructions are estimated from single views.

\begin{figure}[t]
	\centering
	\includegraphics[width=88mm]{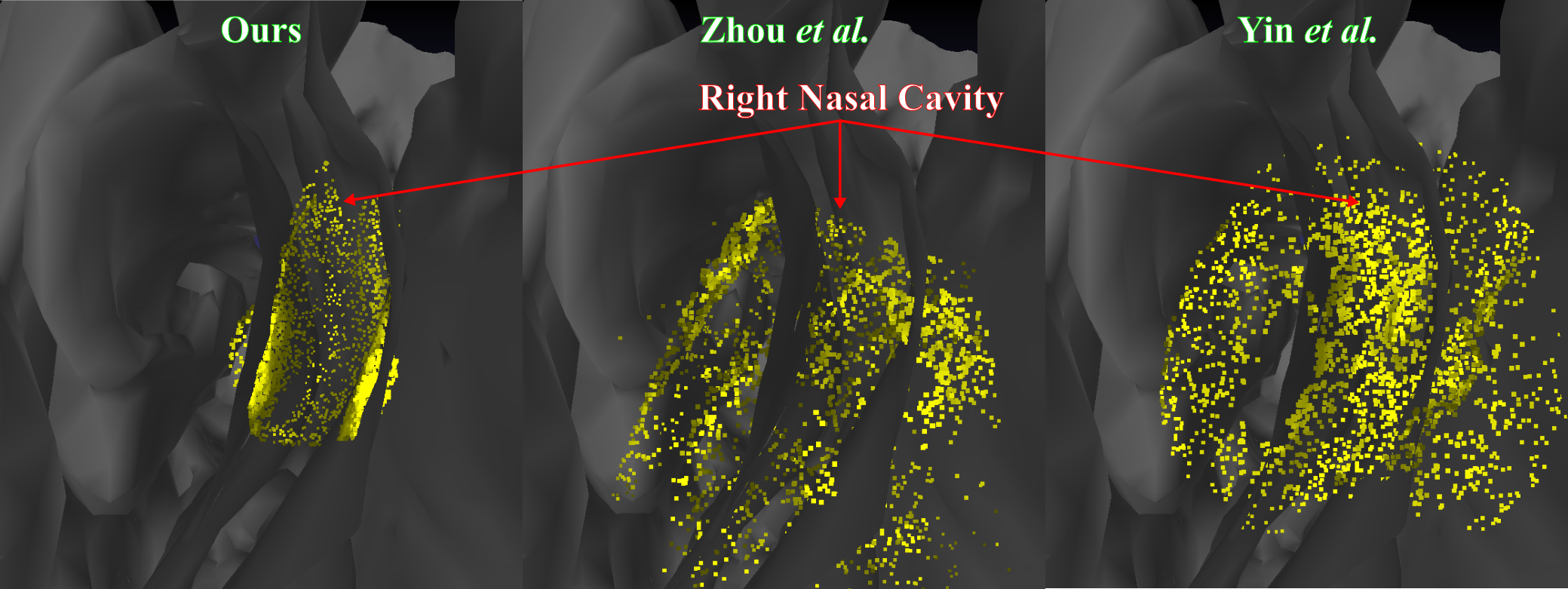}
	\caption{
	    \textbf{Reconstructions registered to patient CT.}\quad Alignment produced between our reconstruction and the corresponding patient CT (left) shows that our reconstruction adheres well to the contours of the patient CT and contains few outliers. Whereas alignment between the reconstructions from Zhou~\etal (middle) and Yin~\etal (right) for the same frame and the corresponding patient CT shows poor alignment and many outliers. Many points of the reconstructions by Zhou~\etal and Yin~\etal fall inside the regions where the endoscope cannot enter.
	    }
	    
	\label{fig:point_cloud_overlay}
\end{figure}

\subsection{Comparison Study}\label{sec:comparisonstudy}
We conduct a comparison study to evaluate the performance of our method against two typical self-supervised depth estimation methods~\cite{zhou2017unsupervised,yin2018geonet}. We use the original implementation of both methods with a slight modification, where we omit the black invalid regions of endoscopy images when computing losses during training. In Fig.~\ref{fig:result_comparison}, we show representative qualitative results for all three methods. In Fig.~\ref{fig:point_cloud_overlay}, we overlay the CT surface model with the registered point clouds of one video frame from all three methods. We also compare our method with these methods quantitatively. Table.~\ref{tab:sfm_result_agreement}, where the evaluation related to SfM is used, shows evaluation results of depth predictions from all three methods, revealing that our method outperforms both competing approaches by a large margin. Note that all video frames from Patient $2$, $3$, $4$, and $5$ are used for evaluation. For this evaluation, all four trained models in the Cross-patient Study are used to generate depth predictions for each corresponding testing patient to test the performance of our method. For Zhou~\etal and Yin~\etal, the evaluation model sees all patient data except Patient $4$ during training. Therefore, it is a comparison in favor of the competing methods. The bad performance of the competing methods on the training and testing dataset shows that it is not overfitting that makes the model performance worse than ours. Instead, these two methods cannot make the network exploit signals in the unlabeled endoscopy data effectively. The boxplot in Fig.~\ref{fig:boxplots} (b) shows the comparison results with the CT surface models. For the ease of experiments, only the data from Patient $4$ are used for this evaluation. The average residual error of our reconstructions is $0.38 \left(\pm 0.13\right) \text{mm}$. For Zhou~\etal, it is $1.77 \left(\pm 1.19\right) \text{mm}$. For Yin~\etal, it is $0.94 \left(\pm 0.36\right) \text{mm}$. The extreme outliers of reconstructions from Zhou~\etal are removed before error calculation.

We believe the main reason for the inferior performance of the two comparison methods lies in the choice of main driving power to achieve self-supervised depth estimation. Zhou~\etal choose L$1$ loss to enforce photometric consistency between two frames. This assumes the appearance of a region does not change when the viewpoint changes, which is not the case in monocular endoscopy where the lighting source moves jointly with the camera. Yin~\etal use a weighted average of Structural Similarity (SSIM) loss and L$1$ loss. SSIM is less susceptible to brightness changes and pays attention to textural differences. However, since only simple statistics of an image patch are used to represent the texture in SSIM, the expressivity is not enough for cases with scarce and homogeneous texture, such as sinus endoscopy and colonoscopy, to avoid bad local minimal during training. This is especially true for the tissue walls present in the sinus endoscopy, where we observe erroneous depth predictions.

\begin{table}
\begin{adjustbox}{max width=88mm}
\begin{threeparttable}
\caption{Evaluation with SfM results\textsuperscript{*}}
\label{tab:sfm_result_agreement}
\centering

\begin{tabular}{ccccc}

\hline
\multirow{2}{*}{Method} & \multirow{2}{*}{Absolute rel. diff.} & \multicolumn{3}{c}{Threshold} \\ \cline{3-5} 
            &                   &$\sigma=1.25$&$\sigma=1.25^2$& $\sigma=1.25^3$\\ \hline
Ours        &$\mathbf{0.20}$&$\mathbf{0.75}$&$\mathbf{0.93}$&$\mathbf{0.98}$ \\
Zhou~\etal~\cite{zhou2017unsupervised}  &$0.66$&$0.41$&$0.68$&$0.83$\\
Yin~\etal~\cite{yin2018geonet}   &$0.41$&$0.54$&$0.78$&$0.89$\\ \hline

\end{tabular}

\begin{tablenotes}
\item[*] The model performance on data from Patient $2$, $3$, $4$, and $5$ is evaluated with two metrics, which are Absolute Relative Difference and Threshold~\cite{yin2018geonet}. The sparse depth maps generated from SfM results are used as groundtruth. The models of our method for evaluation are those used in the cross-patient study, which means the data from all four patients are not seen during training. On the other hand, the models of Zhou~\etal and Yin~\etal have seen data from Patient $2$, $3$, and $5$ during training.
\end{tablenotes}

\end{threeparttable}
\end{adjustbox}
\end{table}

\subsection{Ablation Study}
\begin{figure}[t]
	\centering
	\includegraphics[width=88mm]{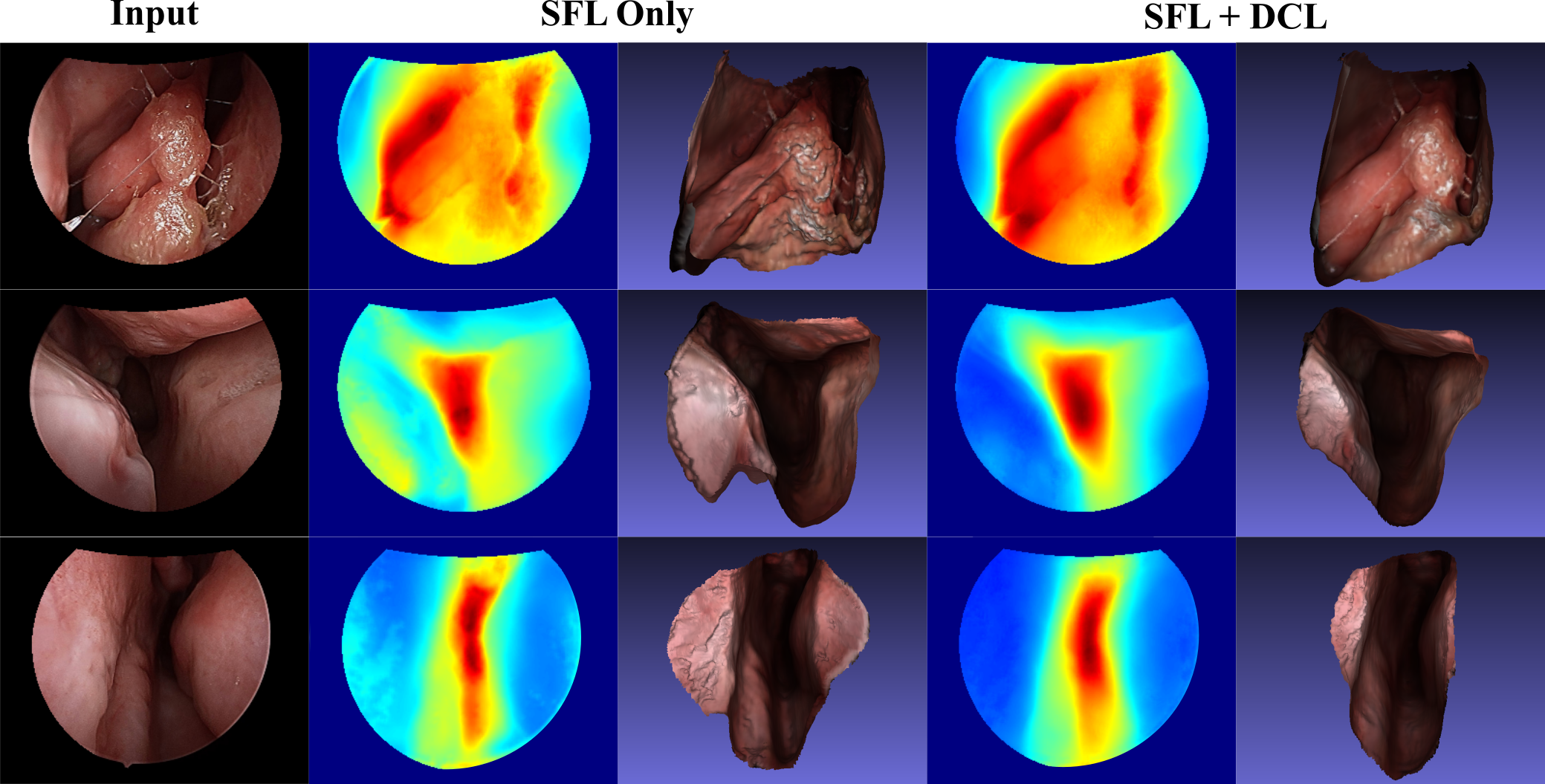} 
	\caption{
	    \textbf{Qualitative result for ablation study.}\quad The results consist of training and testing images, where the first $2$ images are seen during training. The second and third columns consist of corresponding depth maps and reconstructions from the model trained with only SFL. The fourth and fifth columns are from the model trained with both SFL and DCL. The result shows that DCL helps with both training and testing cases. It provides additional guidance to regions where sparse reconstructions from SfM are either inaccurate, \eg~regions with specularity in the first row, or missing, \eg~regions near the boundary in the second and third row.
	}
	\label{fig:ablation_result}
\end{figure}

To evaluate the effect of loss components, \ie~SFL and DCL, a network is trained with only SFL with Patient $4$ for testing. The model trained in the Cross-patient Study with Patient $4$ for testing is used for comparison. Since DCL alone is not able to train a model with meaningful results, we do not evaluate its performance alone. The qualitative (Fig.~\ref{fig:ablation_result}) and quantitative (Fig.~\ref{fig:boxplots} (b)) results show that the model trained with SFL and DCL combined has a better performance than the model trained with SFL only. In terms of the evaluation results on data from Patient $4$, the average residual error for the model trained with SFL only is $0.47 \left(\pm 0.10\right) \text{mm}$. In terms of the evaluation related to SfM, the values of metrics including absolute relative difference, threshold test with $\sigma=1.25, 1.25^2, 1.25^3$ are $0.14$, $0.81$, $0.98$, $1.00$, respectively. In comparison, the average residual error for the model trained with SFL and DCL is $0.38 \left(\pm 0.13\right) \text{mm}$. The values of the same metrics as above are $0.13$, $0.85$, $0.98$, $1.00$, respectively, which shows slight improvement compared with the model trained with SFL only. Note that sparse depth maps are unevenly distributed and there are usually few valid points for evaluation on the tissue wall which DCL is observed to help most with. Therefore, the observed improvement in the evaluation related to SfM is not as large as the average residual error in the evaluation related to CT data.

\section{Discussion} \label{sec:discussion}
The proposed method does not require any labeled data for training and generalizes well across endoscopes and patients. The method was initially designed for and evaluated on sinus endoscopy data, however, we are confident that it is also applicable to monocular endoscopy of other anatomies. However, some limitations of our method remain that need to be addressed in the future work. First, the training phase of our method relies on the reconstructions and camera poses from SfM. On the one hand, this means our method will evolve and improve with more advanced SfM algorithms becoming available. On the other hand, this means our method does not apply to cases where the SfM is not able to produce reasonable results. Whereas our method tolerates random errors and outliers from SfM to a certain extent, if large systematic errors occur in a large portion of the data, which could occur in cases of highly dynamic environments, our method will likely fail. Second, our method only produces dense depth maps up to a global scale. In scenarios where the global scale is required, additional information needs to be provided during the application phase to recover the global scale. This can be achieved \eg~by measuring known-size objects or using external tracking devices. In terms of the inter-frame geometric constraints, concurrent to our work, 3D ICP loss was proposed by~\cite{mahjourian2018unsupervised} to enforce geometric consistency of two depth predictions. Because the Iterative Closest Point (ICP) used in their loss calculation is not differentiable, they use the residual errors of the point cloud registration upon convergence as the difference approximation of two depth predictions. There are two advantages of the proposed DCL over the 3D ICP loss. First, it is able to handle errors between two depth predictions that can be compensated by a rigid transformation. Second, it does not involve a registration method which can potentially introduce erroneous information for training when a registration failure happens. Because the implementation of the 3D ICP loss is not released, no comparison is made in this work. Recently, a similar geometric consistency loss~\cite{bian2019unsupervised} has been proposed, which is subsequent to our work~\cite{liu2018self}. In terms of the evaluation, the average residual error reported in the evaluation related to CT data can lead to underestimated errors. This is because the residual error is calculated using pairs of closest points between the registered point clouds and the CT surface models. Since the distance between a closest point pair is always less than or equal to the distance between the true point pair, the overall error will be underestimated. Depending on the accuracy of SfM, the evaluation related to SfM may better represent the true accuracy for regions of the depth predictions that have valid correspondences in the sparse depth maps. But this metric has the disadvantage that regions where no valid correspondences exist in the sparse depth maps are not evaluated. The exact accuracy estimate is available only if the camera trajectory of a video is accurately registered to the CT surface model, which is what we currently do not have and will work on as a future direction.

\section{Conclusion} \label{sec:conclusion}
In this work, we present a self-supervised approach to training convolutional neural networks for dense depth estimation in monocular endoscopy without any \emph{a priori} modeling of anatomy or shading. To the best of our knowledge, this is the \emph{first} deep learning-based self-supervised depth estimation method proposed for monocular endoscopy. Our method only requires monocular endoscopic videos and a multi-view stereo method during the training phase. In contrast to most competing methods for self-supervised depth estimation, our method does not assume photometric constancy, making it applicable to endoscopy. In a cross-patient study, we demonstrate that our method generalizes well to different patients, achieving submillimeter residual errors even when trained on small amounts of unlabeled training data from several other patients. In a comparison study, we show that our method outperforms two recent self-supervised depth estimation methods by a large margin on \emph{in vivo} sinus endoscopy data. For future work, we plan to fuse depth maps from single frames to form an entire 3D model to make it more suitable for applications such as clinical anatomical study and surgical navigation.

\bibliographystyle{IEEEtran}

\end{document}